\documentclass[12pt,a4paper]{article}
\usepackage[margin=2.5cm]{geometry}
\usepackage{setspace}
\usepackage{microtype}
\usepackage{graphicx}
\usepackage{float}
\usepackage{caption}
\usepackage{subcaption}
\usepackage{booktabs}
\usepackage{multirow}
\usepackage{threeparttable}
\usepackage{array}
\usepackage{makecell}
\usepackage{amsmath,amssymb}
\usepackage{bm}
\usepackage{mathtools}
\usepackage{siunitx}
\usepackage{chemformula}
\usepackage{newtxtext,newtxmath}
\usepackage[super,sort&compress]{natbib}
\usepackage{xcolor}
\usepackage[colorlinks=true,allcolors=blue]{hyperref}
\usepackage{parskip}
\usepackage{titlesec}
\titleformat{\section}{\LARGE\bfseries}{}{1em}{}
\titleformat{\subsection}{\Large\bfseries}{}{1em}{}
\titleformat{\subsubsection}{\large\bfseries}{}{1em}{}

\title{Thermodynamics-Informed Learning Captures Hierarchical Hydrogen-Bond Networks in Deep Eutectic Solvents}
\author{
Yanpeng Ye$^{1}$, Chaoran Duan$^{2}$, Mian Xu$^{1,*}$, S. Mani Sarathy$^{1}$\\[0.7em]
\parbox{0.88\textwidth}{\centering
\small
$^{1}$Clean Energy Research Platform, Physical Science and Engineering Division (PSE),\\
King Abdullah University of Science and Technology, Thuwal 23955-6900, Saudi Arabia;\\[0.4em]
$^{2}$Key Laboratory of the Three Gorges Reservoir Region's Eco-environments, Ministry of Education, Institute of Environment and Ecology, Chongqing University, Chongqing 400045, China.\\[0.6em]
$^{*}$Corresponding author: mian.xu@kaust.edu.sa
}
}
\date{}

\begin{document}
\maketitle

\begin{abstract}
\noindent Deep eutectic solvents (DESs) offer a highly tunable platform for lignocellulose refining, yet their solvation behavior arises from interactions among hydrogen bond acceptors (HBAs), hydrogen bond donors (HBDs), and solutes within multicomponent hydrogen bond networks, rather than from the additive properties of individual constituents. Existing machine learning models represent DES components and operating conditions independently, overlooking the hierarchical and thermodynamic nature of multicomponent solvation and limiting their reliability for new DES formulations and sparse conditions. This work develops TiNN, a thermodynamics-informed neural network that captures hierarchical hydrogen bond networks, from the association between HBAs and HBDs to solute accommodation, and links them to solubility under varying conditions through a learned free energy pathway. Trained on 938 records, TiNN achieves an $R^{2}$ of 0.917 on the test set, with robust generalization to new HBA and HBD combinations and to sparse conditions. Ablation shows that solute-specific stabilization makes the largest contribution to prediction accuracy, while the contribution of hydrogen bond compatibility depends on the dissolution mechanism: it is pronounced for carbohydrate and lignin solutes but negligible for gases, which dissolve through other interactions. In new measurements of glucose, xylan, and Kraft lignin in acidic, basic, and neutral DESs, TiNN reduces the geometric mean fold error of solubility from more than 23 with COSMO-RS to about 1.5. Without training on fractionation data, the same molecular solvation information also separates DESs with high and low removal of lignin and hemicellulose from corn stover, linking the solubility of isolated biomass components to the fractionation of real lignocellulose. These results identify hierarchical hydrogen bonding and solute-specific stabilization as key molecular determinants of DES solvation for biomass-derived solutes, and provide a chemically grounded route to select DESs for lignocellulose fractionation before experiments.
\end{abstract}

\section{Introduction}

Lignocellulose biorefineries serve as a key platform for converting abundant and renewable bioresources into fuels, chemicals, and functional materials, thereby enabling a transition to a circular bioeconomy.\cite{ragauskas2014lignin} The organic components of lignocellulose biomass mainly consist of cellulose, hemicellulose, and lignin, which are chemically and structurally distinct biopolymers that are intrinsically interconnected to form a recalcitrant supramolecular network.\cite{himmel2007biomass} This pronounced heterogeneity in physicochemical properties and chemical reactivities substantially impedes the selective component valorization and often results in harsh processing conditions and side product formation.\cite{rinaldi2016paving} Fractionating lignocellulose into relatively pure individual organic components represents a prevailing trend and essential prerequisite for component-oriented biorefineries, enabling efficient production of biofuels, biochemicals, and biomaterials.\cite{renders2017lignin,li2023hydrogen,xu2024production,xu2024ex} Deep eutectic solvents (DESs) with negligible vapor pressure, favorable biocompatibility, recyclability,\cite{zhang2012deep} and generally inexpensive synthetic materials have emerged as promising alternatives\cite{yue2025lignin, smith2014deep} to conventional concentrated acids, strong alkalis, and organic solvents on lignocellulose fractionation.  The use of these harsh substance results in equipment corrosion, chemical waste generation, and solvent emissions.\cite{francisco2012new,alvarez2016unique,shen2019facile}

DESs can be simply synthesized by mechanically mixing a wide variety of hydrogen bond donors (HBDs) and hydrogen bond acceptors (HBAs) with adjustable molar ratios,\cite{abbott2003novel,abbott2004deep} enabling highly tunable polarity, acidity, and hydrogen-bonding basicity.\cite{zhang2025fluorinated,hansen2020deep,pandey2013polar} This strong hydrogen bond network (HBN) formed between HBDs and HBAs can facilitate the deconstruction and selective dissolution of lignocellulosic components through a reaction-assisted solvation mechanism\cite{chen2018ultrafast}: the HBN dynamically reorganizes to disrupt noncovalent interactions within the biomass matrix, while chemical reactions may also cleave lignin--carbohydrate linkages; DES components competitively replace native hydrogen bonds, van der Waals forces, and aromatic interactions; newly formed component-DES hydrogen bonds, together with ion-dipole interactions, dispersion forces, and $\pi$–$\pi$ interactions, compensate for the loss of polymer-polymer cohesive interactions, thereby stabilizing the liberated molecules in the liquid phase.\cite{ashworth2016doubly} Therefore, the multilevel hydrogen bond network constitutes a key molecular determinant of the lignocellulose fractionation performance of DESs.\cite{hammond2017effect,chen2018ultrafast} However, the compositional tunability of DESs creates a highly complex and reconfigurable hydrogen bond network such that changes in combinations and stoichiometry of HBDs and HBAs ($>$ 100 types) concurrently alter network polarity, acidity, density, and viscosity.  This leads to nonlinear trade-offs between DES-DES association and competitive DES-biomass solvation.\cite{florindo2014insights} Further coupled with operating conditions and feedstock heterogeneity, this tunability makes lignocellulose fractionation performance difficult to infer directly from the bulk property of individual DES constituent. Consequently, the rational design and formulation of optimal DES for lignocellulose refining is a challenging multidimensional screening problem.\cite{lemaoui2023predicting,tan2026dual}  

Conventionally, the screening of DESs for lignocellulose refining has primarily relied on two approaches, i.e., experimental evaluation and quantum chemical solvation modeling. Although experimental evaluation remains the most reliable route for assessing actual fractionation performance, systematically exploring the enormous space of HBA-HBD-solute combinations and operating parameters is prohibitively time- and resource-intensive.\cite{tan2020deep, zhou2022lignin, mqoni2026deep} Quantum chemical solvation models, such as the conductor-like screening model for real solvents (COSMO-RS), can provide useful molecular information on solute-solvent affinity and preliminarily guide solvent screening.\cite{al2025accurately,hansen2020deep, wojeicchowski2020using} However, their accuracy is often constrained by simplifying assumptions and highly sensitive to molecular conformations and may require extensive computational resources when applied to large and diverse candidate spaces.\cite{wang2024assessment, song2021evaluation}

Data-driven methods have therefore attracted increasing attention for the rapid screening of DESs, as they can learn complex and nonlinear composition-structure-property relationships from available data and evaluate extensive chemical spaces at lower computational costs.\cite{xue2026three,butler2018machine,ye2024construction,zhao2026automated} Machine learning models, ranging from descriptor-based regressors to graph neural networks, have recently been applied to predict DES properties, including viscosity, density, and melting point.\cite{Ma2025CATMPNN, Yu2022TSTiNet, Lavrinenko2023EutecticTemperatures} Nevertheless, most existing models describe a DES system as a simple collection of discrete constituents, typically by concatenating constituent-wise molecular descriptors or combining independently encoded HBAs and HBDs with fixed molar ratios.\cite{lemaoui2023predicting, wang2021prediction, sharma2025machine} Such coarse-grained representations are chemically incomplete since they fail to explicitly capture the composition-dependent and dynamically reconfigurable multilevel HBN that governs DES formation, solvent-solute interactions, and final lignocellulose fractionation performance. Moreover, these representations treat the HBA, HBD, and solute as static inputs and treat temperature and pressure as ordinary features,\cite{sheikhshoaei2025data, xing2025novel, mohan2024physics} overlooking both the sequential nature of DES fractionation (the HBA--HBD environment forms prior to solute accommodation) and the thermodynamic origin of the solubilization process.\cite{wang2025thermodynamics, mohan2017solubility, chen2025identifying}

This study develops TiNN, a thermodynamics-informed neural network that represents DES fractionation as two coupled chemical steps: formation of the HBA--HBD hydrogen-bonded environment, followed by accommodation of the solute within this environment. These two steps are linked to a learned free-energy pathway, connecting the molecular representation with the temperature- and pressure-dependent solubility response. Trained on 938 experimental records, TiNN achieves higher prediction accuracy than the conventional COSMO-RS method and captures dissolution trends under varying temperatures, pressures, and HBD:HBA combinations. It also demonstrates compositional generalization across HBA--HBD pairings. The predictions are further validated by solubility measurements on DES-solute pairs spanning acidic, basic, and neutral DES environments. By integrating solvation thermodynamics with the multilevel hydrogen bond network, this chemically interpretable representation provides a readily deployable framework for the rational design of DES in lignocellulose refining.

\section{Results and Discussion}

\subsection{Dataset Distribution}

\begin{figure}[!htbp]
    \centering
    \includegraphics[width=\columnwidth]{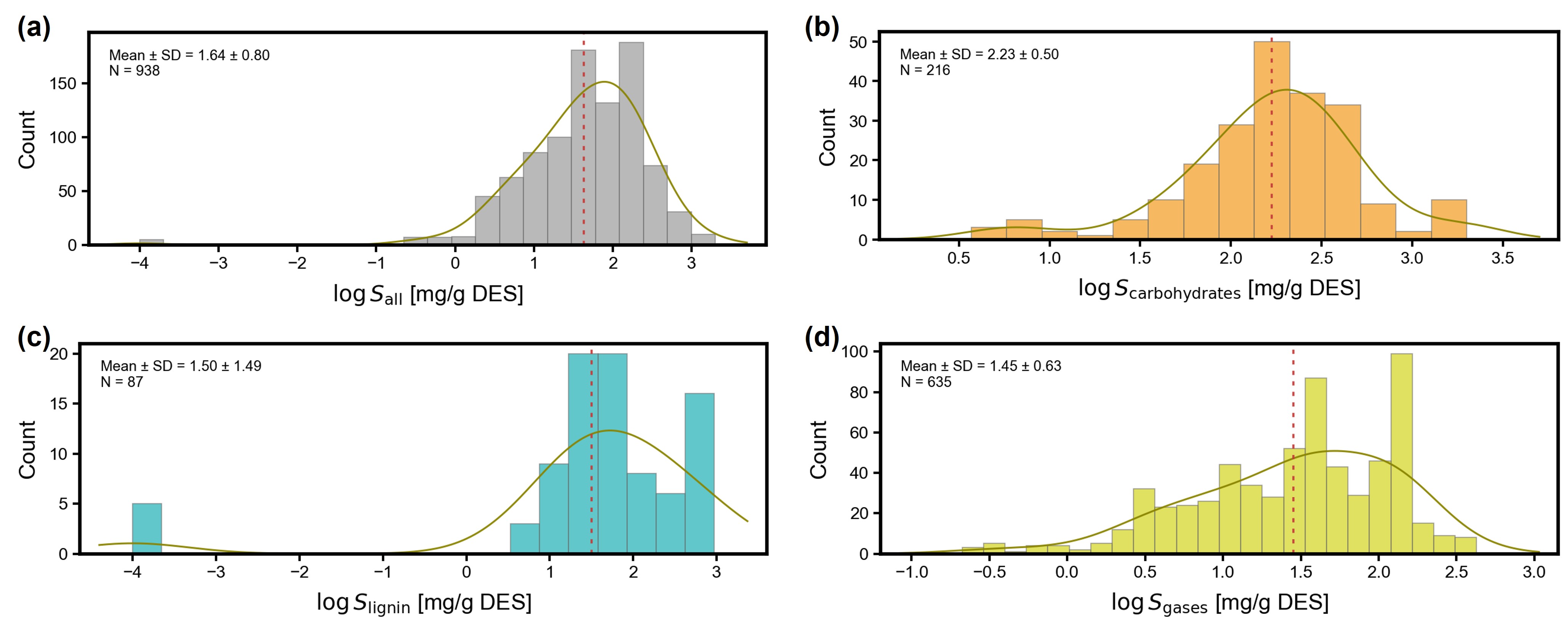}
    \caption{Statistical overview of the DES solubility dataset. Distribution of $\log S$ values for (a) all DES--solute systems, (b) carbohydrates, (c) lignin-derived compounds, and (d) gases. The dashed red lines indicate the mean values.}
    \label{fig:data_statistics}
\end{figure}

The compiled dataset spans a broad range of DES--solute systems and exhibits a strongly heterogeneous solubility distribution (Figure~\ref{fig:data_statistics}). The resulting $\log S$ values have a mean of 1.64 and a standard deviation of 0.80, with substantial variation across different solute families. The solutes were grouped into carbohydrates (cellulose- and hemicellulose-derived compounds), lignin-derived compounds, and gases, providing chemically distinct subsets for evaluating the ability of TiNN to capture different solvation regimes.

\subsection{Thermodynamics-Informed Learning}

\begin{figure}[htbp]
    \centering
    \includegraphics[width=\textwidth]{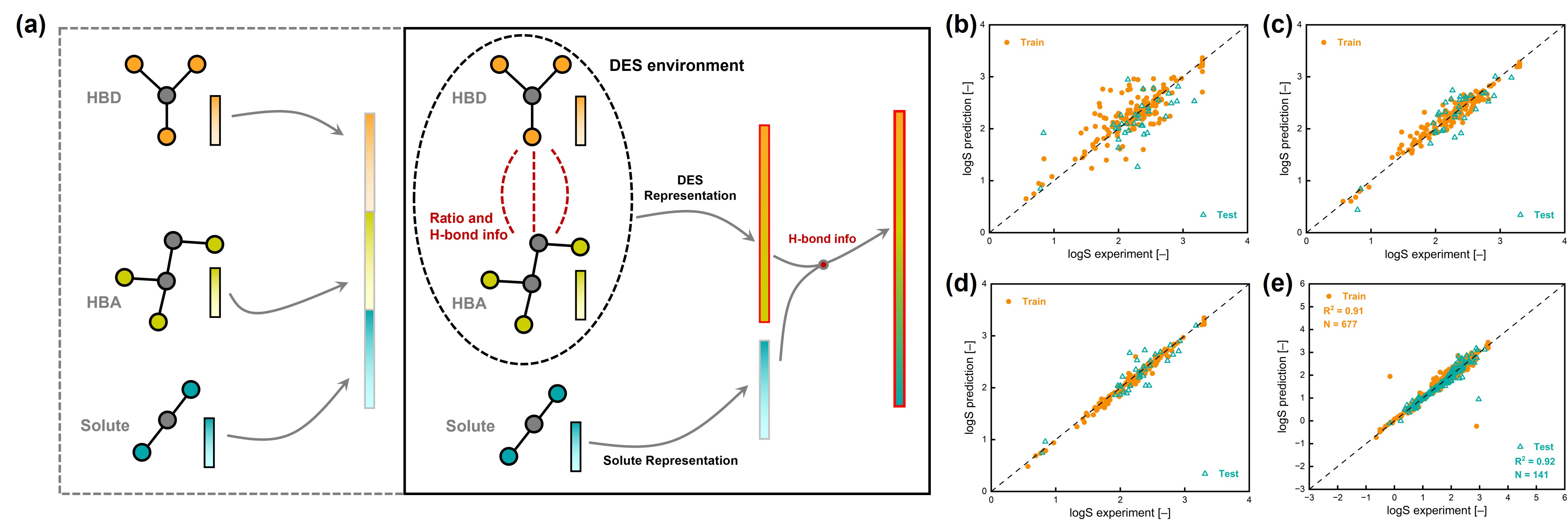}
    \caption{Molecular representation and predictive performance of TiNN. (a) Representation strategies for DES--solute systems. Left: direct concatenation of independently encoded HBA, HBD, and solute embeddings. Right: TiNN first combines the HBA and HBD into a DES representation, incorporating composition and internal hydrogen-bond compatibility, and then integrates the solute through DES--solute hydrogen-bond compatibility. (b--d) Parity plots on the carbohydrate subset for (b) MLP with concatenated HBA, HBD, and solute descriptors, (c) TiNN with concatenated input, and (d) TiNN with the hierarchical three-molecule representation. (e) Training- and test-set predictions of the full TiNN model; the reported test-set metrics are calculated on held-out samples.}
    \label{fig:overview}
\end{figure}

TiNN represents the DES-solute complexes through a structure that accounts for intermolecular interactions and reflects the ordered nature of DES solvation (Figure~\ref{fig:overview}a). In contrast to direct feature concatenation, which treats the HBA, HBD, and solute as independent inputs and combines them at a single level (Figure~\ref{fig:overview}a, left), TiNN first forms a DES-level representation by combining the HBA and HBD embeddings with their molar ratio and the internal hydrogen-bond compatibility, and then couples this DES representation with the solute embedding through the DES-solute hydrogen-bond compatibility (Figure~\ref{fig:overview}a, right). This hierarchical construction encodes the sequence in which the HBA--HBD environment forms before the solute is accommodated. Implementation details of the encoders $\Phi_{\mathrm{DES}}$ and $\Phi_{\mathrm{solv}}$ are given in the Methods section.

The hierarchical representation defines the HBA--HBD--solute complexes, but it does not constrain how solubility changes with temperature by itself. Accordingly, $\log S$ reflects a thermodynamic balance between solute stabilization and the cost of incorporation into the DES liquid phase. In contrast, a direct mapping from the system representation to $\log S$ treats temperature as an ordinary descriptor and leaves this response unconstrained.

TiNN organizes the prediction through a learned free-energy pathway. Two latent quantities, $\Delta H_{\theta}$ and $\Delta S_{\theta}$, are inferred from the system representation. An enthalpy--entropy consistency loss encourages the learned free energy to follow Eq.~\eqref{eq:tinn_delta_g}.
\begin{equation}
\Delta G_{\theta}
\approx
\Delta H_{\theta}(\mathbf{z}_{\mathrm{system}})
-
T\,\Delta S_{\theta}(\mathbf{z}_{\mathrm{system}}).
\label{eq:tinn_delta_g}
\end{equation}
Here, $\Delta H_{\theta}$ and $\Delta S_{\theta}$ are not calibrated to experimental enthalpy and entropy. They provide a thermodynamic inductive bias that separates the learned response into a temperature-independent contribution and a temperature-scaled contribution. The temperature branch supplies the explicit condition input, while the learned free-energy terms constrain how this input affects $\log S$. This provides a thermodynamic regularization of the learned temperature response.

A solute-specific free-energy term is further included to account for the different energetic cost of accommodating different solutes in the same DES environment. Real DES systems can also deviate from the idealized thermodynamic form because of ion association, hydrogen-bond rearrangement, and other non-ideal interactions. TiNN therefore writes the final prediction as
\begin{equation}
\log S_{\mathrm{pred}}
=
\underbrace{\log S_{\mathrm{thermo}}(\mathbf{z}_{\mathrm{system}},T)}_{\text{thermodynamics-guided}}
+
\underbrace{\delta_{\theta}(\mathbf{z}_{\mathrm{system}},T)}_{\text{non-ideal correction}},
\label{eq:tinn_prediction}
\end{equation}
where $\log S_{\mathrm{thermo}}$ describes the thermodynamics-guided response, and $\delta_{\theta}$ captures non-ideal deviations. A thermodynamic consistency constraint is applied to keep the learned temperature response aligned with the free-energy form.

\subsection{Predictive Performance}

\subsubsection{Overall Accuracy of TiNN}

After defining the three-molecule representation and the thermodynamics-informed learning module, the capability of TiNN in predicting DES solubility across chemically diverse DES--solute complexes was investigated. Figure~\ref{fig:overview}e shows the training- and test-set predictions of the full TiNN model. TiNN achieves a test-set $R^{2}$ of 0.917 and maintains close agreement with experimental values over a broad solubility range. A small number of high-error points are observed in the parity plot. This agreement covers cellulose- and hemicellulose-derived carbohydrates, lignin-derived compounds (such as vanillin and syringaldehyde), and gaseous solutes (such as CO$_{2}$, SO$_{2}$, and H$_{2}$S). The broad coverage indicates that TiNN does not only fit a limited group of chemically similar compounds, but can learn solubility trends across distinct solute families.

We then examined how the two main design choices in TiNN, the thermodynamics-informed learning module and the hierarchical three-molecule representation, contributed to this performance. Carbohydrate solutes were selected as a representative subset because they provide sufficient sample size and cover a continuous solubility distribution. As shown in Figure~\ref{fig:overview}b--d, the predictions progressively improve from the MLP baseline to the thermodynamics-informed model with concatenated input, and finally to the full TiNN with the hierarchical three-molecule representation. The improvement from the MLP baseline to the thermodynamics-informed model shows that the learned free-energy pathway helps constrain the solubility response. The further improvement after introducing the hierarchical representation shows that resolving DES formation and DES--solute accommodation provides information that cannot be captured by simple feature aggregation. These results provide an initial indication that DES solubility is better described when both multilevel molecular interaction structure and thermodynamic response are considered, rather than when the system is treated as an additive combination of separated molecular descriptors.

\subsubsection{Comparison with Alternative Models}

To examine whether the two central design ideas of TiNN were the main contributors for its superior performance, they were compared with alternative models at three different levels. First, the 3D molecular encoder within the TiNN framework was changed to examine the effect of geometry encoding. Second, the EGNN-based representation was kept, but the TiNN head was replaced with conventional prediction heads. Third, the TiNN was compared with descriptor-based, graph-based, and literature baselines that do not explicitly represent the ordered HBA--HBD--solute solvation process. The main results are summarized in Table~\ref{tab:baseline_comparison}.

EGNN-TiNN provides the best overall performance, with the lowest RMSE and the highest $R^{2}$ and Spearman correlation. PaiNN-TiNN gives a similar MAE but is slightly weaker in RMSE and $R^{2}$, whereas DimeNet++-TiNN performs substantially worse. This comparison indicates that the three-molecule TiNN framework benefits from a stable 3D molecular encoder. The encoder is therefore not only a technical choice, but affects how effectively the complicated HBA--HBD--solute environment is represented.

The second comparison isolates the role of the thermodynamics-informed prediction structure. EGNN-MLP and EGNN-XGBoost use the same EGNN-based representation as EGNN-TiNN, but both give lower overall performance. This result verifies that the molecular embedding alone is not sufficient. Instead, the learned free-energy pathway provides an additional constraint that helps convert the HBA--HBD--solute representation into a condition-dependent solubility response.

Descriptor and graph baselines provide a further test of the representation itself. XGBoost and CatBoost achieve relatively low MAE and high Spearman correlation, indicating that descriptor models can capture part of the statistical relationship between molecular identity and solubility. However, their RMSE and $R^{2}$ are substantially weaker than those of EGNN-TiNN, showing that they are less effective in reproducing the full quantitative response. GIN and the concatenated SMILES baseline also underperform TiNN, consistent with the limitations of concatenation-based representations discussed above. These comparisons indicate that molecular interactions and thermodynamic structure are both essential for describing DES--solute solubility. Therefore, TiNN is able to address this challenge by combining an ordered HBA--HBD--solute representation with a thermodynamics-informed prediction pathway, allowing the model to treat DES solubility as a multicomponent solvation response rather than as a direct descriptor-to-property mapping.

\begin{table}[htbp]
\centering
\caption{Comparison of TiNN with alternative models for DES--solute solubility prediction. The models test the effects of 3D molecular encoding, thermodynamics-informed prediction, and descriptor-based representations.}
\label{tab:baseline_comparison}
\begin{tabular}{llcccc}
\toprule
Group & Model & MAE & RMSE & $R^{2}$ & Spearman \\
\midrule
\multirow{3}{*}{\makecell[l]{Hierarchical \\ TiNN variants}}
 & EGNN-TiNN          & 0.126           & \textbf{0.242} & \textbf{0.917} & \textbf{0.956} \\
 & PaiNN-TiNN         & \textbf{0.123}  & 0.253          & 0.909          & 0.954 \\
 & DimeNet++-TiNN     & 0.163           & 0.665          & 0.562          & 0.886 \\
\midrule
\multirow{2}{*}{\makecell[l]{EGNN representation \\ + conventional heads}}
 & EGNN-MLP           & 0.170           & 0.286          & 0.884          & 0.913 \\
 & EGNN-XGBoost       & 0.145           & 0.493          & 0.656          & 0.935 \\
\midrule
\multirow{5}{*}{\makecell[l]{Descriptor and \\ graph baselines}}
 & MLP                & 0.232           & 0.595          & 0.498          & 0.878 \\
 & XGBoost            & 0.127           & 0.489          & 0.661          & 0.929 \\
 & CatBoost           & 0.138           & 0.426          & 0.743          & 0.938 \\
 & GIN                & 0.301           & 0.604          & 0.484          & 0.808 \\
 & Concat-SMILES Ridge       & 0.259           & 0.485          & 0.667          & 0.848 \\
\bottomrule
\end{tabular}
\end{table}

\subsubsection{Ablation Study}

To further identify which design can promote the performance of TiNN, the full model was compared with a series of simplified variants trained under the same setup. These variants tested two parts of TiNN. For the molecular representation, the 3D molecular encoder was replaced by 2D descriptors of comparable dimensionality. For the thermodynamics-informed learning module, variants were tested without the residual branch, the solute-specific free-energy term, the thermodynamic consistency constraint, or the temperature branch (Table~\ref{tab:ablation}).

The largest performance drop occurs when the solute-specific free-energy term is removed ($\Delta R^{2} = -0.163$; Table~\ref{tab:ablation}), showing that solute-dependent stabilization cannot be recovered by the shared molecular representation alone. Removing the residual branch gives the second largest drop ($\Delta R^{2} = -0.102$), demonstrating that the conventional ideal-solubility limit is insufficient for real DES complexes and that non-ideal correction is necessary. Removing the temperature branch or the thermodynamic consistency constraint causes a smaller decrease ($\Delta R^{2} = -0.031$ and $-0.028$), suggesting that TiNN learns temperature dependence through both the explicit temperature encoder and the thermodynamic loss. Replacing the 3D molecular representation with 2D descriptors also leads to consistent degradation ($\Delta R^{2} = -0.022$), in agreement with the encoder comparison in Table~\ref{tab:baseline_comparison}.

Overall, the ablation results show that both parts of TiNN are necessary: the 3D representation provides molecular geometry beyond 2D descriptors, while the thermodynamics-informed module separates solute-specific stabilization, thermodynamic consistency, and non-ideal deviations in the solubility response. These results further indicate that DES solubility is better modeled as a thermodynamically constrained HBA--HBD--solute response than as a direct descriptor-to-property mapping.

\begin{table}[htbp]
\centering
\caption{Ablation of TiNN components on the test set. Each variant removes or replaces one component of the full model. $\Delta R^{2}$ is relative to the full TiNN.}
\label{tab:ablation}
\begin{tabular}{lccccc}
\toprule
Variant & MAE & RMSE & $R^{2}$ & Spearman & $\Delta R^{2}$ \\
\midrule
Full TiNN                              & \textbf{0.126} & \textbf{0.242} & \textbf{0.917} & \textbf{0.956} & --       \\
w/o solute-specific free-energy term   & 0.183 & 0.417 & 0.754 & 0.924 & $-0.163$ \\
w/o residual branch                    & 0.173 & 0.361 & 0.815 & 0.937 & $-0.102$ \\
w/o temperature branch                 & 0.149 & 0.284 & 0.886 & 0.941 & $-0.031$ \\
w/o thermodynamic consistency          & 0.151 & 0.280 & 0.889 & 0.941 & $-0.028$ \\
3D encoder $\rightarrow$ 2D descriptors & 0.155 & 0.273 & 0.895 & 0.935 & $-0.022$ \\
\bottomrule
\end{tabular}
\end{table}

\subsection{Robustness and Generalization}

The condition-dependent response and compositional generalization of TiNN were further examined across representative DES--solute systems and held-out HBA--HBD pairings.

\subsubsection{Condition-Dependent Solubility Response}

\begin{figure}[htbp]
    \centering
    \includegraphics[width=\textwidth]{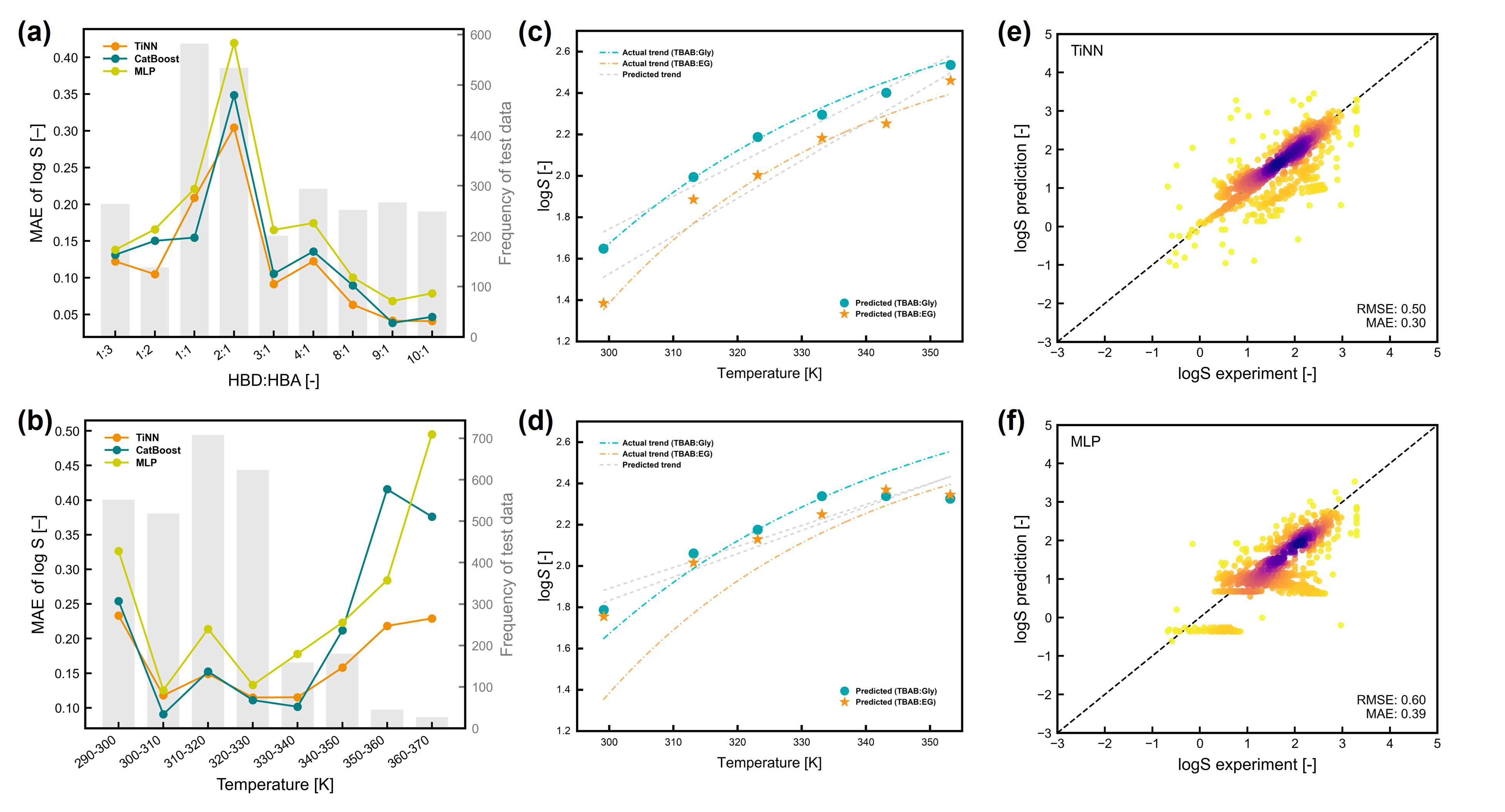}
    \caption{Condition-dependent response and compositional generalization of TiNN. (a, b) MAE of TiNN, CatBoost, and MLP across HBD:HBA ratio bins (a) and temperature bins (b); gray bars indicate the number of test samples in each bin. (c, d) Temperature-dependent SO$_{2}$ solubility in TBAB:Gly and TBAB:EG predicted by TiNN (c) and MLP (d); dash-dotted lines show the experimental trends and gray dashed lines show the predicted trends. (e, f) Out-of-fold parity plots for TiNN (e) and MLP (f) under leave-one-HBA--HBD-pair-out cross-validation. Each of the 81 HBA--HBD combinations was held out once, with all associated records excluded from training. Color indicates point density.}
    \label{fig:condition}
\end{figure}

Moreover, whether TiNN captures the dependence of DES--solute solubility on physical conditions was studied with a focus on the HBD:HBA molar ratio and temperature. Figure~\ref{fig:condition}a,b shows the MAE of TiNN, CatBoost, and MLP across ratio and temperature bins, with the sample count in each bin shown by gray bars. In data-rich regions, the three models show comparable errors. Clearer differences appear near the boundaries of the dataset. Above 340~K, where the number of test samples decreases sharply, CatBoost and MLP show larger error increases, whereas TiNN remains closer to its overall error level.

The predicted and experimental condition-dependent trends were then compared for representative DES--solute systems. Figure~\ref{fig:condition}c,d shows the temperature dependence of SO$_2$ solubility in tetrabutylammonium bromide (TBAB):glycerol (Gly) and TBAB:ethylene glycol (EG) predicted by TiNN and MLP, respectively. TiNN follows both the magnitude and the slope of the experimental trends. In contrast, the MLP predictions are flatter and deviate further from the experimental trend of TBAB:EG.

These results demonstrate that TiNN does not simply interpolate toward an average response over the training distribution but rather retains condition-dependent variation in DES--solute solubility, especially in sparse temperature and ratio regions. This behavior is consistent with the thermodynamics-informed branch, which constrains the temperature-resolved response, and the hierarchical three-molecule representation, which captures system-specific deviations under different pressure and composition conditions.

\subsubsection{Compositional Generalization}

TiNN's compositional generalization was investigated across HBA--HBD combinations within the solubility task.

Compositional generalization was evaluated using leave-one-HBA--HBD-pair-out cross-validation across 81 distinct HBA--HBD combinations. In each fold, all records associated with one HBA--HBD pair were excluded from training and used for testing, including records with different molar ratios, solutes, temperatures, and pressures. The individual HBA or HBD could still occur in other training combinations, so this evaluation specifically tests whether the model can generalize to new pairings of known molecular components.

The aggregated out-of-fold predictions show that TiNN preserves a clear diagonal parity distribution, whereas the MLP baseline develops horizontal bands indicative of regression toward an averaged response (Figure~\ref{fig:condition}e,f). This behavior indicates that the hierarchical HBA--HBD--solute representation captures composition-level relationships that can be transferred across HBA--HBD pairings, rather than relying primarily on memorization of specific combinations.

\subsection{Chemical Significance of Hydrogen-Bond Compatibility}

The remarkable performance and generalization results above show that TiNN predicts DES-solute solubility accurately across chemically diverse systems. Accordingly, the origin of this performance was examined to determine whether it reflects a chemically consistent use of the hydrogen-bond compatibility signal incorporated into the model or whether the encoded compatibility merely functions as an arbitrary numerical anchor. However, it cannot be simply inferred from the predictive accuracy alone: the multilevel hydrogen bond network is the molecular determinant of DES-solute interactions and the chemical motivation behind the hierarchical HBA-HBD-solute representation used in TiNN. Therefore, the role of this hydrogen-bonding structure in model predictions was probed through two analyses of the relative hydrogen-bond compatibility descriptor $H^{\mathrm{rel}}$ defined in Methods: an ablation of the $H^{\mathrm{rel}}$ input channel under three perturbations, and an examination of the $H^{\mathrm{rel}}$ value distribution across solute classes (Figure~\ref{fig:hbn_exp}a--g).

The TiNN was retrained with three modifications of the $H^{\mathrm{rel}}$ input: values kept unchanged (Full), values set to zero (Zero), and values shuffled across DES-solute pairs while preserving the marginal distribution (Shuffled). On the full test set, Full and Zero performed within 0.023 in $R^{2}$, while Shuffled dropped by 0.217 (Figure~\ref{fig:hbn_exp}a). This asymmetry became more informative when the test set was partitioned by solute type. On the carbohydrate (cellulose- and hemicellulose-derived) and lignin-derived subset, where solutes carry O-H donor and acceptor sites, Full outperforms Zero by 0.046 and Shuffled by 0.412 in $R^{2}$. On the gas subset, Full and Zero were indistinguishable ($\Delta R^{2} = -0.001$), consistent with the absence of classical O-H or N-H donor sites on CO$_{2}$, SO$_{2}$, and H$_{2}$S. Rather than relying on classical donor–acceptor hydrogen bonding, gas solutes dissolve in DES through distinct mechanisms: CO$_2$ primarily through cavity formation and dispersion interactions,\cite{pelaquim2021gas, chi2025mechanisms} SO$_2$ through charge-transfer and Lewis acid–base interactions with anionic components,\cite{li2015theoretical} and H$_2$S through Lewis acid–base interactions.\cite{wang2023h2s} Consequently, the contribution of $H^{\mathrm{rel}}$ to TiNN predictions is mechanism-dependent, i.e., it is pronounced when hydrogen-bond compatibility is chemically informative, but negligible when dissolution is dominated by mechanisms not represented by $H^{\mathrm{rel}}$.

The $H^{\mathrm{rel}}$ input carries chemically meaningful patterns across the dataset itself (Figure~\ref{fig:hbn_exp}b--g). $H_{\text{DES--solute}}^{\mathrm{rel}}$ spans more than eight orders of magnitude across solute classes: mostly 1--3 for carbohydrates, which present multiple hydroxyl groups as both donors and acceptors; below 2 for lignin-derived compounds, whose aromatic scaffolds carry fewer donor sites; and $\sim 10^{-8}$ for gaseous solutes, which fall to near zero because CO$_{2}$, SO$_{2}$, and H$_{2}$S provide no classical O-H or N-H donor. $H_{\text{HBA--HBD}}^{\mathrm{rel}}$, which characterizes the internal hydrogen-bond balance of the DES, follows the same rule-based construction. For carbohydrate and lignin-derived solutes, it takes lower values than the DES-solute channel, reflecting the tighter internal donor-acceptor pairing between HBA and HBD relative to the DES-solute interface. For gaseous solutes, the DES-solute channel falls to near zero, so the HBA-HBD channel becomes the larger of the two. These value patterns arise directly from the rule-based donor-acceptor scoring of the constituent molecules and reflect the intrinsic hydrogen-bond structure of the DES-solute system rather than any external labeling or supervision.

The ablation and distribution results show that TiNN incorporates $H^{\mathrm{rel}}$ as a chemically meaningful signal rather than an arbitrary numerical channel. The model responds to the chemical pattern of $H^{\mathrm{rel}}$ values, preferentially uses this signal where hydrogen-bond compatibility governs solvation, and disregards it where dissolution proceeds through other mechanisms. Combined with the ordered HBA-HBD-solute representation and the thermodynamically constrained temperature response established above, this behavior further indicates that TiNN captures DES-solute solubility as a chemically interpretable manifestation of multicomponent solvation, rather than as a direct descriptor-to-property mapping.

\subsection{Solubility and Biomass Fractionation Experiment}

\begin{figure}[htbp]
\centering
\includegraphics[width=\textwidth]{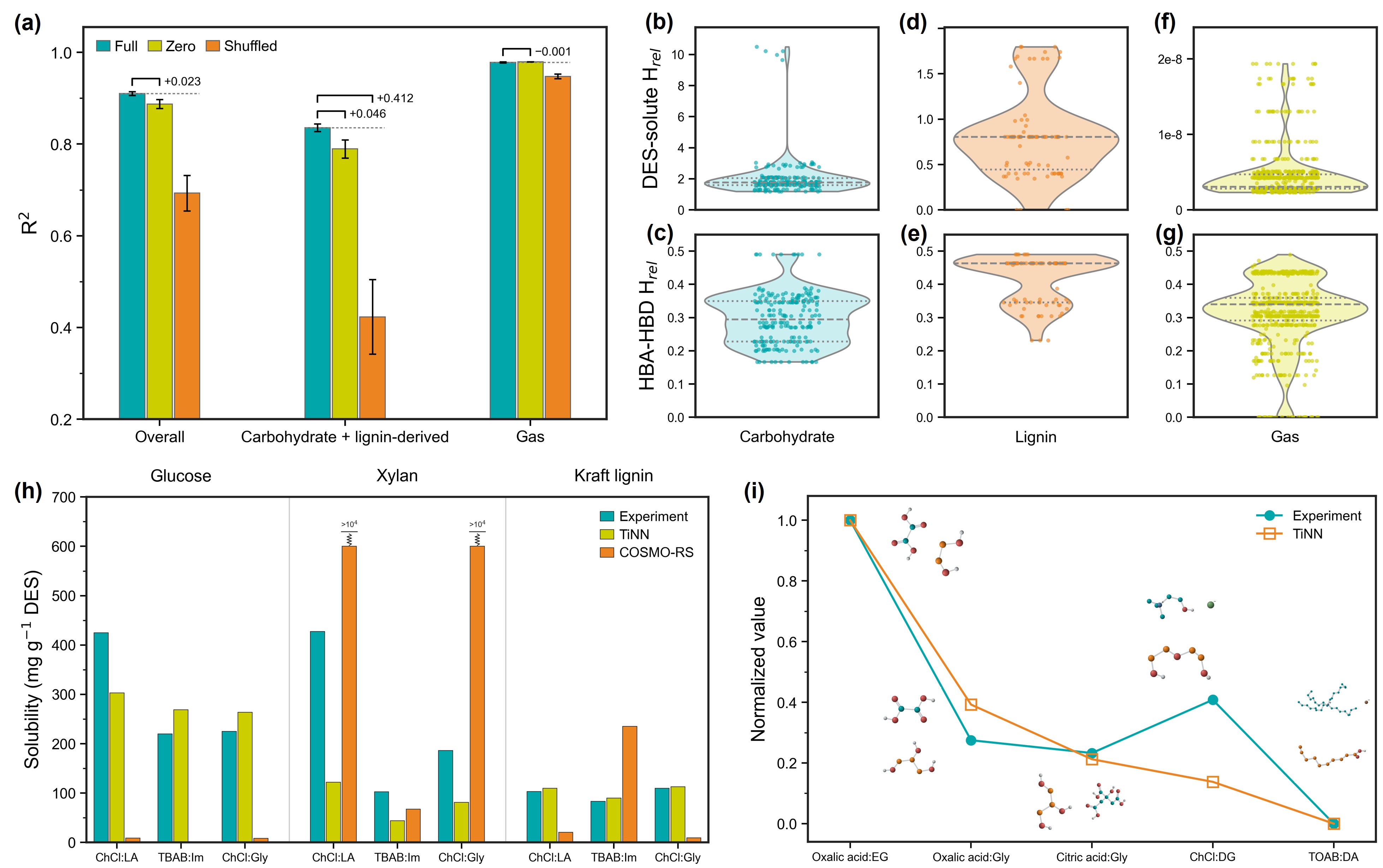}
\caption{Hydrogen-bond compatibility and experimental validation of TiNN. (a) Ablation of the $H^{\mathrm{rel}}$ input under three perturbations; bracket labels give $R^{2}(\text{Full}) - R^{2}(\text{variant})$ and error bars indicate one standard deviation over three splits. (b--g) Distributions of $H_{\text{DES--solute}}^{\mathrm{rel}}$ (b, d, f) and $H_{\text{HBA--HBD}}^{\mathrm{rel}}$ (c, e, g) for carbohydrate, lignin-derived, and gaseous solutes. (h) Measured and predicted solubilities in three DESs at 323.15~K; broken bars mark COSMO-RS predictions of complete miscibility. (i) Normalized TiNN predictions and corn stover fractionation for five DESs selected across the predicted dissolution ranking.}
\label{fig:hbn_exp}
\end{figure}

The predictive capability of TiNN was first experimentally examined using three representative DES environments: acidic ChCl:lactic acid (ChCl:LA), basic TBAB:imidazole (TBAB:Im), and relatively neutral ChCl:glycerol (ChCl:Gly). Their solubilization performance for cellulose-derived glucose, hemicellulose-derived xylan, and Kraft lignin was investigated at 323.15~K and 101.325~kPa with an HBD:HBA molar ratio of 2:1. The experimental values were compared with predictions from TiNN and the conventional COSMO-RS method (Figure~\ref{fig:hbn_exp}h).

Across the nine DES--solute systems, TiNN achieves a $\log_{10}$-scale MAE of 0.18, compared with at least 1.37 for COSMO-RS, corresponding to geometric mean fold errors of approximately 1.5 and at least 23, respectively. For the two systems predicted by COSMO-RS to be completely miscible, a conservative surrogate solubility of $10^{4}$~mg~g$^{-1}$ DES was assigned solely for error calculation. TiNN shows close agreement with the experimental measurements for glucose and Kraft lignin and correctly reproduces the experimental ranking of xylan solubility across the three DESs, although its absolute values are underestimated. This deviation may partly arise from the polydisperse nature of xylan,\cite{scheller2010hemicelluloses} which is represented by an idealized molecular structure in the model, together with the relatively limited xylan data available for training.

To further examine whether the molecular solubility predictions can support practical DES screening, five DESs spanning different predicted lignin--xylan dissolution capacities were evaluated using real lignocellulose (corn stover). The normalized TiNN predictions were compared with the experimentally measured dissolution of lignin and hemicellulose (Figure~\ref{fig:hbn_exp}i). Although the intermediate DESs are not reproduced in their exact order, TiNN correctly distinguishes the DESs with the highest and lowest fractionation capacities and captures the overall experimental trend. This result indicates that the HBA--HBD--solute interaction patterns learned by TiNN retain practical relevance in lignocellulose fractionation, where disruption and reorganization of the hydrogen-bond network contributes to lignin and hemicellulose dissolution.

Overall, the two levels of experimental validation show that TiNN not only reproduces molecular solubility trends across chemically distinct DES environments, but can also provide a useful screening signal for distinguishing DESs with relatively high and low lignocellulose fractionation capability. Importantly, the model is not directly trained on whole-biomass fractionation data, and the agreement in Figure~\ref{fig:hbn_exp}i therefore reflects transfer of the learned molecular solvation information to a more complex real biomass system.

\section{Conclusions}

The present work established a thermodynamics-informed framework that resolves DES--solute solvation through hierarchical hydrogen-bond networks and structured multicomponent interactions, enabling molecularly informed DES design for lignocellulose refining. Instead of treating the solvation as a direct descriptor-to-property mapping, the developed TiNN resolves the formation of the HBA--HBD environment and the subsequent accommodation of the solute, while constraining the condition-dependent response through a learned free-energy pathway. This chemically structured representation achieves a test-set $R^{2}$ of 0.917 and shows condition-dependent behavior and generalization across held-out HBA--HBD pairings. The mechanism-dependent contribution of hydrogen-bond compatibility further supports this interpretation: shuffling this signal decreases $R^{2}$ by 0.412 for carbohydrate and lignin-derived solutes, whereas removing it produces essentially no change for the gas subset.

Furthermore, the experimental results substantiate the remarkable transferability of TiNN. Across independently measured glucose, xylan, and Kraft lignin solubilities in acidic, basic, and neutral DES environments, TiNN achieves a geometric mean error of approximately 1.5-fold, compared with more than 23-fold for COSMO-RS, indicating that the learned HBA--HBD--solute relationships extend beyond the specific combinations represented in the training data. More notably, this molecular solvation information remains informative when transferred to the substantially more complex process of real lignocellulose fractionation. Although TiNN was not trained on fractionation data and does not explicitly represent the structural and kinetic complexity of lignocellulose deconstruction, its predictions still distinguish DES formulations with relatively high and low lignin--hemicellulose fractionation performances. Overall, this cross-scale transfer suggests that the chemically structured solvation learning can provide a meaningful bridge between molecular property prediction and practical solvent screening, offering a route toward more physically informed and rational design of multicomponent DES systems.

\section{Methods}

\subsection{Dataset and Model Representation}

The dataset contains 938 DES--solute solubility records, each specifying the HBA, HBD, solute, temperature, pressure, and HBD:HBA ratio. Solubility was expressed in mg~g$^{-1}$ DES and modeled as $\log S=\log_{10}(S)$. Solute-stratified splitting assigned approximately 72\%, 13\%, and 15\% of records to training, validation, and testing, respectively.

The HBA, HBD, and solute were encoded separately from COSMO-RS molecular conformations using an EGNN. Their embeddings were combined hierarchically, first into a DES representation and then into a DES--solute representation:
\begin{equation}
\begin{aligned}
\mathbf{z}_{\mathrm{DES}} &= \Phi_{\mathrm{DES}}(x_{\mathrm{HBA}}\mathbf{z}_{\mathrm{HBA}},x_{\mathrm{HBD}}\mathbf{z}_{\mathrm{HBD}},H_{\text{HBA--HBD}}^{\mathrm{rel}}),\\
\mathbf{z}_{\mathrm{system}} &= \Phi_{\mathrm{solv}}(\mathbf{z}_{\mathrm{DES}},\mathbf{z}_{\mathrm{solute}},H_{\text{DES--solute}}^{\mathrm{rel}}),
\end{aligned}
\end{equation}
where $x$ denotes molar fraction. Hydrogen-bond donor and acceptor scores ($D_i$ and $A_i$) were assigned using rule-based molecular-site identification with local chemical corrections. The relative DES--solute compatibility was defined as
\begin{equation}
H^{\mathrm{rel}}=\frac{\min(D_{\mathrm{DES}}A_{\mathrm{solute}},A_{\mathrm{DES}}D_{\mathrm{solute}})+\epsilon}{A_{\mathrm{DES}}+D_{\mathrm{DES}}+\epsilon},
\label{eq:H_rel}
\end{equation}
with DES-level capacities calculated as molar-fraction-weighted sums. The same form was applied to the HBA--HBD pair.

\subsection{Thermodynamics-Informed Learning}

The latent free energy combined solute-specific, DES--solute coupling, and temperature-dependent contributions:
\begin{equation}
\Delta G_{\theta}=\Delta G_{\theta}^{\mathrm{solute}}(\mathbf{z}_{\mathrm{solute}})+\Delta G_{\theta}^{\mathrm{env}}(\mathbf{z}_{\mathrm{DES}},\mathbf{z}_{\mathrm{solute}})+\Delta G_{\theta}^{T}(\mathbf{z}_{\mathrm{system}},\mathbf{z}_{T}),
\label{eq:dg_decomp}
\end{equation}
where $\mathbf{z}_{T}=\Phi_T(T)$. The enthalpy--entropy relation in Eq.~\eqref{eq:tinn_delta_g} and the relation $\log S_{\mathrm{thermo}}+\Delta G_{\theta}/(\alpha T)\approx0$ were imposed as soft constraints, with trainable scale $\alpha$. A residual term was added as in Eq.~\eqref{eq:tinn_prediction}. Training jointly minimized log-solubility prediction error and thermodynamic regularization; the checkpoint with the highest validation $R^2$ was selected.

Performance was assessed using MAE, RMSE, $R^2$, and Spearman correlation. Compositional generalization was evaluated by holding out each of 81 distinct HBA--HBD pairs in turn, regardless of molar ratio. For each fold, the other pairs were split into training (85\%) and validation (15\%) sets, and held-out predictions were aggregated.

\subsection{Experimental Validation}

Glucose, xylan, and Kraft lignin solubilities were measured in ChCl:LA, TBAB:Im, and ChCl:Gly (HBD:HBA = 2:1) at 323.15~K. Solute was added incrementally until undissolved particles were observed; solubility was reported as $S=1000m_s/m_{\mathrm{DES}}$ (mg~g$^{-1}$ DES), with $m_s$ and $m_{\mathrm{DES}}$ in grams.

Five DESs selected across the predicted xylan--Kraft lignin dissolution ranking were tested for corn stover fractionation at 373.15~K using a biomass-to-DES mass ratio of 1:40. Cellulose, hemicellulose, and lignin contents were compared before and after treatment.

\bibliography{references}

\end{document}